# Subjective and Objective Visual Quality Assessment of Textured 3D Meshes

JINJIANG GUO and VINCENT VIDAL, Univ Lyon, CNRS, LIRIS, France
IRENE CHENG and ANUP BASU, Multimedia Research Center, University of Alberta, Canada
ATILLA BASKURT and GUILLAUME LAVOUE, Univ Lyon, CNRS, LIRIS, France

Objective visual quality assessment of 3D models is a fundamental issue in computer graphics. Quality assessment metrics may allow a wide range of processes to be guided and evaluated, such as level of detail creation, compression, filtering, and so on. Most computer graphics assets are composed of geometric surfaces on which several texture images can be mapped to make the rendering more realistic. While some quality assessment metrics exist for geometric surfaces, almost no research has been conducted on the evaluation of texture-mapped 3D models. In this context, we present a new subjective study to evaluate the perceptual quality of textured meshes, based on a paired comparison protocol. We introduce both texture and geometry distortions on a set of 5 reference models to produce a database of 136 distorted models, evaluated using two rendering protocols. Based on analysis of the results, we propose two new metrics for visual quality assessment of textured mesh, as optimized linear combinations of accurate geometry and texture quality measurements. These proposed perceptual metrics outperform their counterparts in terms of correlation with human opinion. The database, along with the associated subjective scores, will be made publicly available online.

## 1. INTRODUCTION

Texture mapped 3D graphics are now commonplace in many applications, including digital entertainment, cultural heritage, and architecture. They consist of geometric surfaces on which several texture images can be mapped to make the rendering more realistic. Common texture maps include the diffuse map, the normal map, and the specular map. After their creation (by a designer or using a scanning/reconstruction process), these textured 3D assets may be subject to diverse processing operations, including simplification, compression, filtering, watermarking, and so on. For instance, with the goal of accelerating transmission for remote web-based 3D visualization (e.g., for a virtual museum

Authors' addresses: J. Guo, V. Vidal, A. Baskurt, and G. Lavoué, LIRIS CNRS, INSA Lyon, Bâtiment Jules Verne, 20 Avenue Albert Einstein, 69621 Villeurbanne, France; emails: jinjiang.guo@insa-lyon.fr, {vvidal, atilla.baskurt, glavoue}@liris.cnrs.fr; I. Cheng and A. Basu, Multimedia Research Centre, Department of Computing Science, University of Alberta, 114St. - 89 Av. Edmonton Alberta, Canada T6G 2E8; emails: {locheng, basu}@ualberta.ca.

application), the geometry may be simplified and quantized, and the texture maps may be subject to JPEG compression. Similar geometry and texture degradations may also occur when these assets have to be adapted for lightweight mobile devices; in that case, textures may have to be sub-sampled or compressed using some GPU-friendly random-access methods (e.g., Ström and Akenine-Möller [2005]). These geometry and texture content corruptions may severely impact the visual quality of the 3D data. Therefore, there is a critical need for efficient perceptual metrics to evaluate the visual impact of these textured model artifacts on the visual quality of the rendered image.

Many visual quality metrics have been introduced in the field of computer graphics. However, most of them can only be applied on images created during the rendering step. They mostly focus on detecting artifacts caused by global illumination approximation or tone mapping [Aydın et al. 2010; Herzog et al. 2012; Yeganeh and Wang 2013; Čadík et al. 2013]. On the contrary, another class of method focuses on evaluating the artifacts introduced on the 3D assets themselves. However, most consider only geometric distortions [Lavoué 2011; Váša and Rus 2012; Wang et al. 2012]. Little work has been conducted to evaluate the visual impact of both geometry and texture distortions on the appearance of the rendered image. Studying the complex perceptual interactions between these two types of information requires a ground-truth of subjective opinions on a variety of models with such degradations. To the best of our knowledge, only Pan et al. [2005] conducted such a subjective study. However, they considered only geometry and texture sub-sampling distortions. In this article, we present a large-scale subjective experiment for this purpose, based on a paired comparison protocol. As in Pan et al. [2005], we restrict the texture information to the diffuse maps. Our dataset contains 272 videos of animated 3D models created from five reference objects, five types of distortions, and two rendering settings. The experiments involved more than 100 people. After an analysis of the influence of lighting, as well as shape and texture content on the perception of artifacts, we then use this subjective ground-truth to evaluate the performance of a large set of state-of-the-art metrics (dedicated to image, video, and 3D models). Finally, we propose new metrics based on optimal combinations of geometric and image measurements.

The rest of this article is organized as follows. Section 2 provides a review of related work, while Section 3 describes our subjective experiments and their results. Section 4 presents a comprehensive evaluation of state-of-the-art image and mesh metrics with respect to our subjective ground-truth, along with details on our proposed perceptual metrics and their validation. Finally, concluding remarks and perspective on the work are outlined in Section 5.

## 2. RELATED WORK

Our goal in this work is to propose a visual quality metric for textured 3D models. Since this involves both geometry and 2D image information, we first review related work in image quality assessment and 3D mesh quality assessment. We then focus on the few works that have addressed the visual quality assessment of textured 3D models. For a global view of the topic of visual quality assessment in computer graphics, the reader may refer to Lavoué and Mantiuk [2015].

### 2.1 Visual Quality Assessment of 2D Images

In the field of 2D image processing, research into objective image quality assessment metrics is substantially developed [Wang and Bovik 2006]. Existing algorithms can be classified according to the availability of a reference image: full reference (FR), no-reference (NR), and reduced-reference (RR). The following discussion only focuses on FR methods, where the original *distortion-free* image is known as the reference image. Since the pioneering work of Mannos and Sakrison [1974], many metrics have been introduced with the aim of replacing the classical peak-signal-to-noise ratio (PSNR) which does not correlate well with human vision. Many techniques have tried to mimic the low-level mechanisms of the human visual system (HVS), such as the *contrast sensitivity function* (CSF), usually modeled by a band-pass filter, and the *visual masking* effect which defines the fact that one visual pattern can hide the visibility of another. These bottom-up approaches include the Sarnoff Visual Discrimination model (VDM) [Lubin 1993], the Visible Difference Predictor (VDP) [Daly 1993], and the more recent HDR-VDP-2 [Mantiuk et al. 2011], suited for any range of luminance. These computational metrics mostly focus on the visual detectability of near-threshold distortions and are usually less efficient for quantifying visual fidelity (i.e., supra-threshold distortions), except some works like VSNR [Chandler and Hemami 2007], which explicitly incorporates both models. In contrast to these computationally bottom-up approaches, some authors proposed top-down metrics which do not take into account any HVS models but instead operate based on some intuitive hypotheses of what the HVS attempts to achieve when shown a distorted image. The most well-known example is the Structural SIMilarity index (SSIM) [Wang et al. 2004] and its derivatives (Multi-Scale SSIM [Wang et al. 2003] and Information-weighted SSIM [Wang and Li 2011]). With a more theoretical definition, the Visual Information Fidelity (VIF) metric [Sheikh and Bovik 2006] was developed with the aim of quantifying the loss of image information resulting from the distortion process. Recent surveys and benchmarks [Zhang 2012] point out the superiority of these top-down approaches for visual fidelity (or quality) prediction. It is interesting, however, to note that these evaluations have been conducted on natural image databases. For the particular case of computer-generated images, Čadík et al. [2012] have shown that distortions introduced by common global illumination approximations are not adequately captured by these popular image metrics (e.g., SSIM).

## 2.2 Visual Quality Assessment of 3D Meshes

Inspired by image quality metrics, several perceptually motivated metrics have been designed for 3D meshes. They are all *full-reference* and attempt to predict the visual fidelity of a given 3D mesh (subject to various geometric distortions) with respect to a reference one. The first authors who tried to incorporate some perceptual insights to improve the reliability of geometric distortion measurements were Karni and Gotsman [2000], who proposed combining the Root Mean Square (RMS) distance between corresponding vertices with the RMS distance of their Laplacian coordinates (which reflect a degree of smoothness of the surface). Lavoué [2011] and Torkhani et al. [2012] proposed metrics based on local differences of curvature statistics, while Váša and Rus [2012] considered the dihedral angle differences. These metrics consider local variations of attribute values at vertex or edge level, which are then pooled into a global score. In contrast, Corsini et al. [2007] and Wang et al. [2012] compute global roughness values per model and then derive a simple global roughness difference. Similar to bottom-up image quality metrics, some of these latter algorithms [Torkhani et al. 2012; Wang et al. 2012; Váša and Rus 2012] integrate perceptually motivated mechanisms, such as visual masking. A recent survey [Corsini et al. 2013] details these works and compares their performance with respect to their correlation with mean opinion scores derived from subjective rating experiments. This study shows that MSDM2 [Lavoué 2011], FMPD [Wang et al. 2012], and DAME [Váša and Rus 2012] are excellent predictors of visual quality. Besides these works on global visual fidelity assessment (suited for supra-threshold distortions), several relevant works were introduced very recently: Nader et al. [2016] introduced a bottom-up visibility threshold predictor for 3D meshes (assuming a flat-shaded rendering), and Guo et al. [2015] studied the *local* visibility of geometric artifacts and showed that curvature may be a good predictor of local distortions. Finally, a comprehensive study was introduced by Lavoué et al. [2016] to investigate the use of image metrics computed on rendered images for assessing the visual quality of 3D models (without texture). It shows that some of them (Multi-Scale SSIM (MS-SSIM), in particular) may offer excellent performance. One of the contributions of the present work, is to validate whether this conclusion still holds for textured meshes.

2.3 Visual Quality Assessment of Textured 3D Models

Only a few publications can be found in the literature dealing with quality assessment of textured 3D models. Existing works [Tian and AlRegib 2004; Yang et al. 2004; Pan et al. 2005] are mostly dedicated to choosing the appropriate mesh and texture levels of detail (LoD) for optimizing progressive transmission of textured meshes. Pan et al. [2005] introduced a metric directly based on mesh and texture resolutions, fitted on subjective data. Even though this metric is efficient for the purpose of optimizing transmission [Cheng and Basu 2007], it cannot be generalized to other types of distortions. Tian and AlRegib [2004, 2008] proposed the Fast Quality Measure (FQM) as a weighted combination of two simple error measurements: the mean squared surface error, approximated in the 2D domain for efficiency issues, and the mean squared error over texture pixels. The weighting coefficient between these two measurements is computed by studying the pixel error on rendered images of several LoDs (considered as a meaningful prediction of their subjective perceptual quality). The FQM metric was not subject to any perceptual validation by a subjective experiment; thus, in the experimental section, we compare our results with it. Yang et al. [2004] and Griffin and Olano [2015] also considered image metrics computed on rendered images as *oracles* of subjective quality. Yang et al. [2004] used the mean squared error over pixels and the Mannos model [Mannos and Sakrison 1974] for optimizing textured mesh transmission, while Griffin and Olano [2015] considered SSIM [Wang et al. 2004] to evaluate the masking effect between texture and normal maps (in the context of compression artifacts). As stated in the above section, our subjective data will provide a means of testing this hypothesis of performance of image metrics computed on rendered images or videos for textured mesh quality assessment.

Finally, two related works are those by Ferwerda et al. [1997] and Qu and Meyer [2008]. The authors aim to evaluate how texture is able to mask geometric distortions (from simplification and remeshing), whereas our objective is to evaluate the effect of texture distortions (possibly combined with geometric distortions) on the final appearance.

3. SUBJECTIVE EXPERIMENT

We conducted a large-scale subjective experiment to evaluate the visual impact of texture and geometry distortions on the appearance of textured 3D models. We chose a paired comparison technique, where observers are shown two stimuli side by side and are asked to choose the one that is most similar to the reference (forced-choice methodology). This protocol was shown to be more accurate than others (e.g., single stimulus rating) due to the simplicity of the subjects' task [Mantiuk et al. 2012]. This section provides details on the subjective study and its results.

3.1 Stimuli Generation

We selected five textured triangle meshes created using different methodologies and targeting different application domains (see Figure 1). The *Hulk* and *Sport Car* are artificial models created using a modeling software. Selected from a community model repository (ShareGC.com), they both have structured texture content and smooth texture seams (i.e., vertices associated with multiple texture coordinate pairs). The *squirrel* and the *Easter Island statue* come from a reconstruction process using multiple photographs, and are courtesy of the EPFL Computer Graphics and Geometry Laboratory. Finally, the *Dwarf* is a scanned model, courtesy of the ISTI-CNR Visual Computing Laboratory, Pisa (http://vcg.isti.cnr.it). These last three models, created respectively from reconstruction and scanning, exhibit a noisier texture image, and two of them have very complex texture seams.

The number of vertices of the five models ranges from 6,000 to 250,000, while the texture size ranges from $256 \times 256$ to $4,096 \times 4,096$. Note that the *Hulk*, *Statue*, and *Sport Car* are associated with several texture images. The full characteristics are detailed in Table I (wireframes with texture seams are illustrated in the supplementary material). These objects span a wide variety of geometry and texture data.

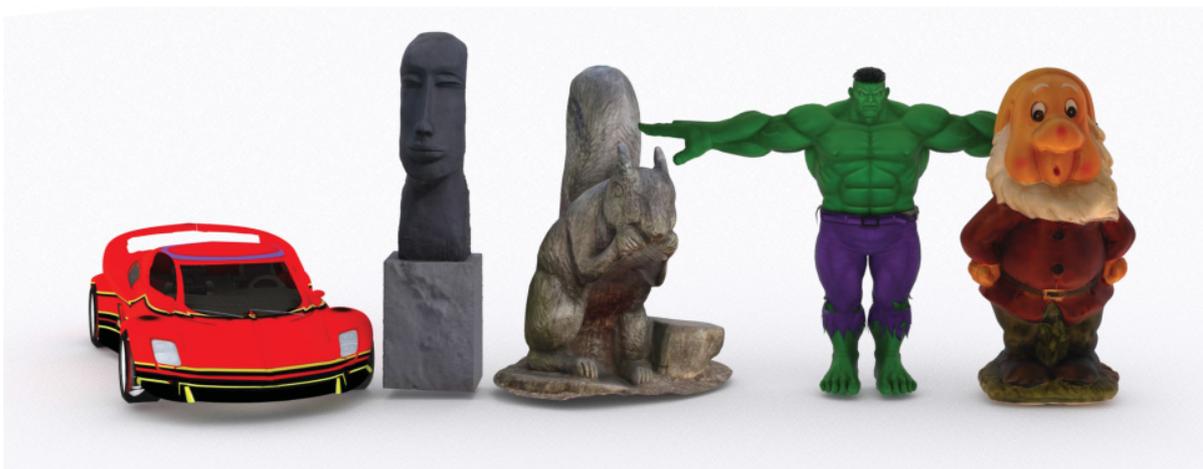

Fig. 1. Five models used in the subjective study (from left to right: Sport Car, Easter Island statue, squirrel, Hulk, and Dwarf).

Table I. Details on our 3D Models: Vertex Number, Texture Image Number, Sizes of Texture Images, Average Shape Curvature, Characteristics of the Texture Images, and Complexity of the Mapping (i.e., of Texture Seams)

|  | #Vertex | #Texture | Texture Size | Average Curv. | Text. Charac. | Map. Complex. |
|---|---|---|---|---|---|---|
| **Squirrel** | 6,185 | 1 | 2048×2048 | High | High freq. & Noisy | Simple |
| **Hulk** | 10,236 | 2 | 1,024×1,024 & 512×512 | High | Structured | Simple |
| **Statue** | 104,019 | 4 | 355×226 to 4,096×4,096 | Low | Noisy | Complex |
| **Sport Car** | 122,873 | 15 | 256×256 to 1,024×768 | Low & sharp edges | Structured | Simple |
| **Dwarf** | 250,004 | 1 | 4,096×4,096 | Intermediate | Intermediate | Complex |

These reference models have been corrupted by five types of distortions (three applied on the geometry and two applied on the texture), each applied with four different strengths:

**On the geometry**:

—*Compression:* We consider uniform geometric quantization, the most common lossy process of compression algorithms.
—*Simplification:* We consider the *Quadric Error Metric* algorithm by Garland and Heckbert [1997].
—*Smoothing*: We consider Laplacian smoothing [Taubin 1995].

**On the texture map**:

—*JPEG*: The most commonly used algorithm for lossy 2D image compression.
—*Sub-sampling*: We reduce texture size by resampling through bilinear interpolation.

The strength of these distortions was adjusted manually in order to span the whole range of visual quality from imperceptible levels to high levels of impairment. For this task, a large set of distortions was generated and viewed by the authors, and a sub-set of them spanning the desired visual quality (i.e., "Excellent," "Good," "Fair," and "Poor") was chosen to be included in the database. This perceptual adjustment of distortion strength was also carried out for the LIVE Video Quality Database [Seshadrinathan et al. 2010]. Thus, we generated 100 distorted models (5 distortion types × 4 strengths × 5 reference models). Table II provides details on the distortion parameters, while Figure 2 illustrates some visual examples.

Table II. Details on the Distortions Applied to Each Reference Model

| ID | Distortion type | Squirrel | Hulk | Statue | Sport Car | Dwarf |
|---|---|---|---|---|---|---|
| L1 | Smoothing | 1 iteration | 1 iteration | 10 iterations | 1 iteration | 15 iterations |
| L2 | Smoothing | 3 iterations | 2 iterations | 20 iterations | 2 iterations | 25 iterations |
| L3 | Smoothing | 5 iterations | 3 iterations | 30 iterations | 3 iterations | 40 iterations |
| L4 | Smoothing | 7 iterations | 4 iterations | 50 iterations | 4 iterations | 50 iterations |
| Si1 | Simplification | 50% removed | 30% removed | 50% removed | 50% removed | 80% removed |
| Si2 | Simplification | 70% removed | 40% removed | 70% removed | 60% removed | 92% removed |
| Si3 | Simplification | 75% removed | 50% removed | 87.5% removed | 75% removed | 97.5% removed |
| Si4 | Simplification | 87.5% removed | 70% removed | 95% removed | 87.5% removed | 98.7% removed |
| Q1 | Quantization | 10 bits | 10 bits | 10 bits | 10 bits | 11 bits |
| Q2 | Quantization | 9 bits | 9 bits | 9 bits | 9 bits | 10 bits |
| Q3 | Quantization | 8 bits | 8 bits | 8 bits | 8 bits | 9 bits |
| Q4 | Quantization | 7 bits | 7 bits | 7 bits | 7 bits | 8 bits |
| J1 | JPEG | 18% quality | 18% quality | 80% quality | 10% quality | 12% quality |
| J2 | JPEG | 14% quality | 14% quality | 16% quality | 5% quality | 10% quality |
| J3 | JPEG | 10% quality | 10% quality | 12% quality | 3% quality | 8% quality |
| J4 | JPEG | 6% quality | 8% quality | 8% quality | 1% quality | 6% quality |
| Su1 | Sub-sampling | 40% sampled | 40% sampled | 25% sampled | 50% sampled | 10% sampled |
| Su2 | Sub-sampling | 30% sampled | 30% sampled | 20% sampled | 20% sampled | 8% sampled |
| Su3 | Sub-sampling | 20% sampled | 20% sampled | 10% sampled | 10% sampled | 5% sampled |
| Su4 | Sub-sampling | 10% sampled | 10% sampled | 5% sampled | 5% sampled | 3% sampled |

### 3.2 Rendering Parameters

**User interaction**: In existing subjective studies involving 3D content, different ways have been used to display the 3D models to the observers, from the most simple (such as static images, as in Watson et al. [2001]) to the most complex (by allowing free rotation, zoom, and translation, as in Corsini et al. [2007]). While it is important for the observer to have access to different viewpoints of the 3D object, the problem of allowing free interaction is the cognitive overload which may alter the results. A good compromise is to use animations, as in Pan et al. [2005]. For each object in our database, we generate a low-speed rotation animation around the vertical axis.

**Lighting and shading**: As noticed by Rogowitz and Rushmeier [2001], the position and type of light sources have a strong influence on the perception of the artifacts. Lighting from the front tends to have a masking effect; hence, we chose an indirect illumination. Sun and Perona [1998] showed that people tend to assume light is above and slightly to the left of the object when they interpret a shaded picture as a 3D scene. Their observations have been confirmed by O'Shea et al. [2008], who demonstrated that the viewer's perception of a 3D shape is more accurate when the angle between the light direction and viewing direction is 20 to 30 degrees above the viewpoint and to the left by 12 degrees from vertical. We follow this lighting condition by putting a spot light at this position. For the material, we kept the original parameters from the source reference objects, which are mainly diffuse. The video resolution is 1920 × 1080. The duration of the video is 15 seconds.

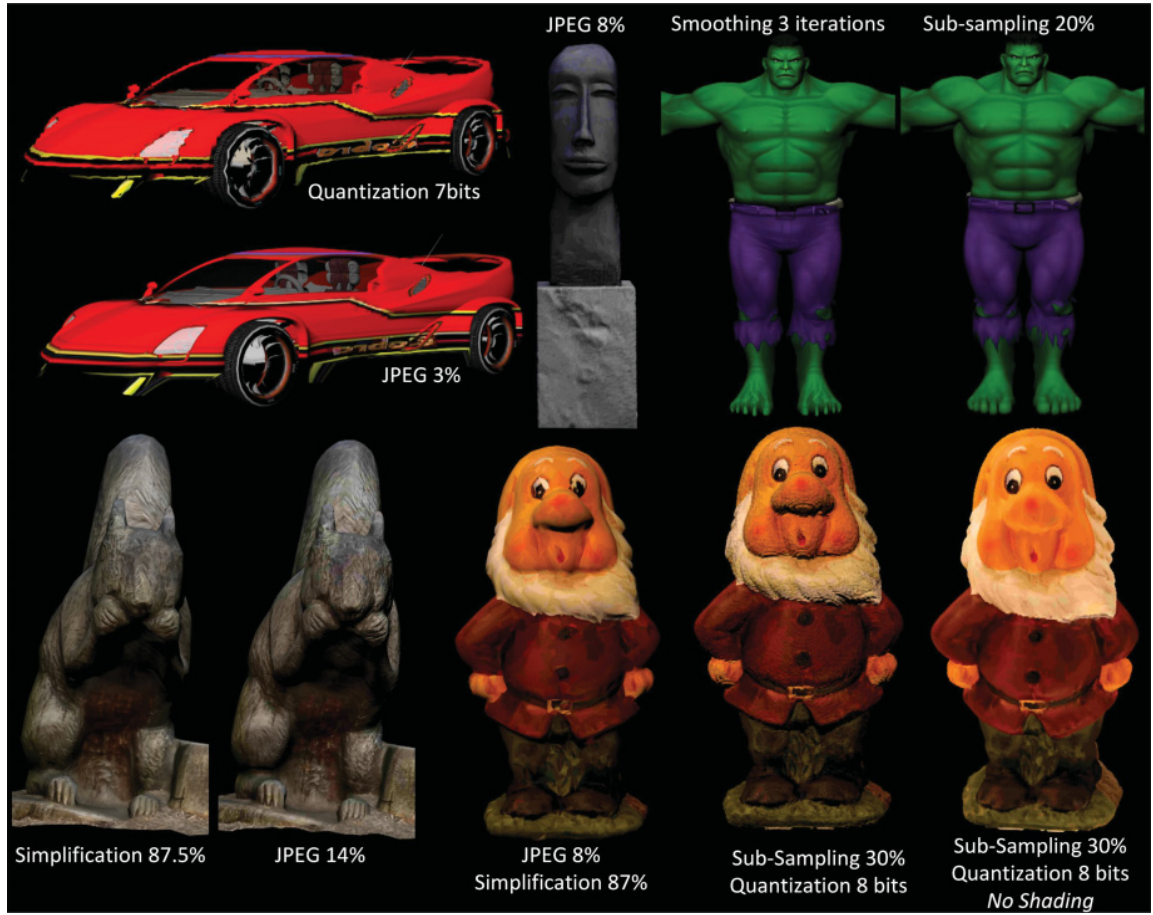

Fig. 2. Examples of distorted models from our dataset. The rendering is the same as in our videos. Note that for the Dwarf, we present here compound distortions (geometry and texture) from our validation set (see Section 4.5).

In order to study the influence of shading on the results, we also re-generated the same set of videos by keeping only the diffuse albedo (i.e., without shading). An example of such rendering is shown in Figure 2, for the Dwarf. We thus obtain 10 video sets (5 models × 2 rendering settings), for a total of 200 videos to evaluate.

3.3 Experimental Procedure

As stated above, we opted for a paired comparison methodology, since this has been demonstrated to be more reliable than rating methods [Mantiuk et al. 2012]. Participants are shown two videos of distorted models at a time, side by side, and are asked to choose the one that is most similar to the reference. The observer can replay the videos as many times as (s)he wants. For the sake of readability, the reference video is not displayed on the same screen but is presented just before the beginning of the comparison, and can then be viewed at any time on a pop-up window by clicking on a button. The interface was developed in JavaScript, as a web platform, and is illustrated in Figure 3.

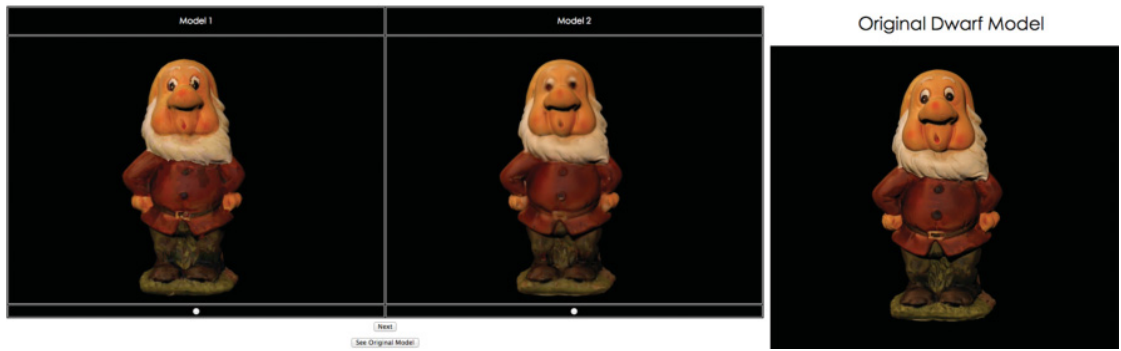

Fig. 3. Illustration of our browser-based interface for the paired comparison task. Pairs of 3D models are presented as videos; the user selects the one (s)he prefers and then clicks on the *Next* button. The reference model can be displayed at any time by clicking on the *See Original Model* button below the *Next* button.

The main issue with the paired comparison protocol is the large number of possible comparisons: $\binom{2}{20} = 190$ per video set. It is therefore unrealistic to ask a participant to perform a complete test even on a single video set. Fortunately, this large number of trials can be reduced by using sorting algorithms as recommended in Silverstein and Farrell [2001] and Mantiuk et al. [2012]. The idea is to embed a sorting algorithm into the experimental platform; this algorithm then decides in an on-line fashion which pairs of videos to compare based on the previous comparisons.

We introduced a simple yet efficient sorting algorithm. The idea is to obtain a global ranking of the 20 stimuli by interleaving them progressively, one distortion type at a time (i.e., compression, simplification, smoothing, JPEG, and sub-sampling). For a given distortion $D$, we assume that the distortion strength ranges from $D_1$ (weak) to $D_4$ (strong). The assumption behind our algorithm is that for a given distortion $D$, the quality of $D_i$ is always better than $D_j$ for $j > i$ (noted as $Q_{D_i} > Q_{D_j}, \forall j > i$). First, two distortion types $Q$ and $J$ are randomly chosen (in this example, quantization and JPEG), $Q_4$ and $J_4$ are then compared. The index of the *not-selected* video (e.g., $Q_4$) is pushed into a list (*List 1*), as the poorest quality version. In the next trial, the *selected* model from the previous round ($J_4$) and a distorted model with a *decreased* level from the other type ($Q_3$) are shuffled and displayed to the user. This process continues until all eight models are sorted from the worst to the best quality. This sorting process is repeated with two other distortion types (i.e., smoothing and simplification) to form a second list (*List 2*) which is then interleaved with the remaining distortion type (sub-sampling) and then with *List 1* to obtain the final ranking of the 20 distortions. In our study, the average comparison number was 36 (instead of 190 for the full design). This sorting algorithm is illustrated in the supplementary material.

### 3.4 Participants

A total of 101 subjects took part in the experiment, aged between 20 and 55, all with normal or corrected to normal vision. Participants were students and staff from the University of Lyon in France and the University of Alberta in Canada. On average, it took 12 minutes for one observer to finish the experiment for one set of videos. Eighty-nine subjects rated one set, eight rated two sets, and four rated three sets. None of these repetitions took place on the same day in order to prevent any learning effect. The experiments were all conducted on the same 15-inch MacBook Pro screen, in a dark room. In total, each of the 10 video sets (5 models × 2 rendering settings) was judged by between 11 and 15 observers (see Table III).

Table III. Agreement between Observers (Kendall's *W* between Their Ranks) for Each Reference Model and Each Rendering Condition

|  | Rendering with shading | | | Rendering without shading | | |
| --- | --- | --- | --- | --- | --- | --- |
|  | Observer Number | Kendall's W | p-value | Observer Number | Kendall's W | p-value |
| **Squirrel** | 11 | 0.69 | <0.0001 | 11 | 0.71 | <0.0001 |
| **Hulk** | 11 | 0.77 | <0.0001 | 13 | 0.70 | <0.0001 |
| **Statue** | 11 | 0.83 | <0.0001 | 15 | 0.76 | <0.0001 |
| **Sport Car** | 11 | 0.74 | <0.0001 | 11 | 0.76 | <0.0001 |
| **Dwarf** | 11 | 0.72 | <0.0001 | 12 | 0.77 | <0.0001 |

### 3.5 Computing Scores

For each video set and each subject, we obtain a global ranking of the $n_m$ distorted models ($n_m$ equals 20 in our experiment). From this ranking, it is easy to retrieve the full preference matrix ($n_m \times n_m$), by applying the transitive relation: if object A is better than object B and B is better than C, then we can deduce that A is better than C. These per-subject preference matrices can then be summed into a single one (per video set). In this matrix $P$, each element $P_{i,j}$ represents the number of times the stimulus $i$ was judged to be of higher quality than stimulus $j$. As in Ledda et al. [2005] and Mantiuk et al. [2012], we then consider the number of votes received by each stimuli as its quality score, which may then be divided by the number of human subjects $n_s$ for normalization among video sets:

$$s_i = \frac{\sum_{j=1}^{n_m} P_{i,j}}{n_s} \quad (1)$$

We thus obtain one subjective score for each distorted model $s_i$ which belongs to [0,19]. Note that 19 occurs only for models which have been ranked first by every subject. Note that more sophisticated statistical methods exist for inferring scale values from a preference matrix. We computed scores using Thurstone's Law of Comparative Judgments, Case V [Thurstone 1927], which assumes that observers' choices can be thought of as sampled from a normal distribution of underlying quality scores. Since the values obtained were very close to the simple vote counts described above (more than 0.99 Pearson correlation between them), we decided to keep the latter.

### 3.6 Analysis and Discussion

3.6.1 *Observer Agreement.* It is essential to analyze the agreements between the subjects before studying the results of the experiment. Since each observer outputs a global ranking of the stimuli, the best way to evaluate their agreement is to compute Kendall's coefficient of concordance $W$ [Govindarajulu et al. 1992] which assesses the agreement among raters. Table III gives details on the results. $W$ ranges from one, meaning complete agreement, to zero, meaning no agreement, while the p-value associated with $W$ provides the likelihood of null hypothesis, which means no agreement between all the subjects. Table III shows that the overall Kendall's $W$ coefficients are at least larger than 0.69, implying a strong agreement among the subjects, confirmed by the very low *p-values*.

3.6.2 *Influence of Shape and Texture on the Visual Impact of the Distortions.* The impact of the distortions greatly depends on the textured 3D models and their characteristics as illustrated in Figure 4. We observe that geometric quantization is less visible on 3D shapes with a few thousands vertices (like the Squirrel) than on shapes with more than 100,000 (like the Dwarf, Sport Car, and Statue). The Hulk model is rather low-resolution but is also strongly damaged by the quantization; the reason is that its head has a high density of triangles and thus is severely damaged by this distortion. The sampling density of the surface actually influences the spatial frequency of the visual distortions created by quantization and thus their perceptual impact due to the Contrast Sensitivity Function (CSF) of the HVS. Note that, while there is a limit on the mesh density, beyond which the distortions become invisible again (as a result of overly high frequency), all of our models are below this threshold.

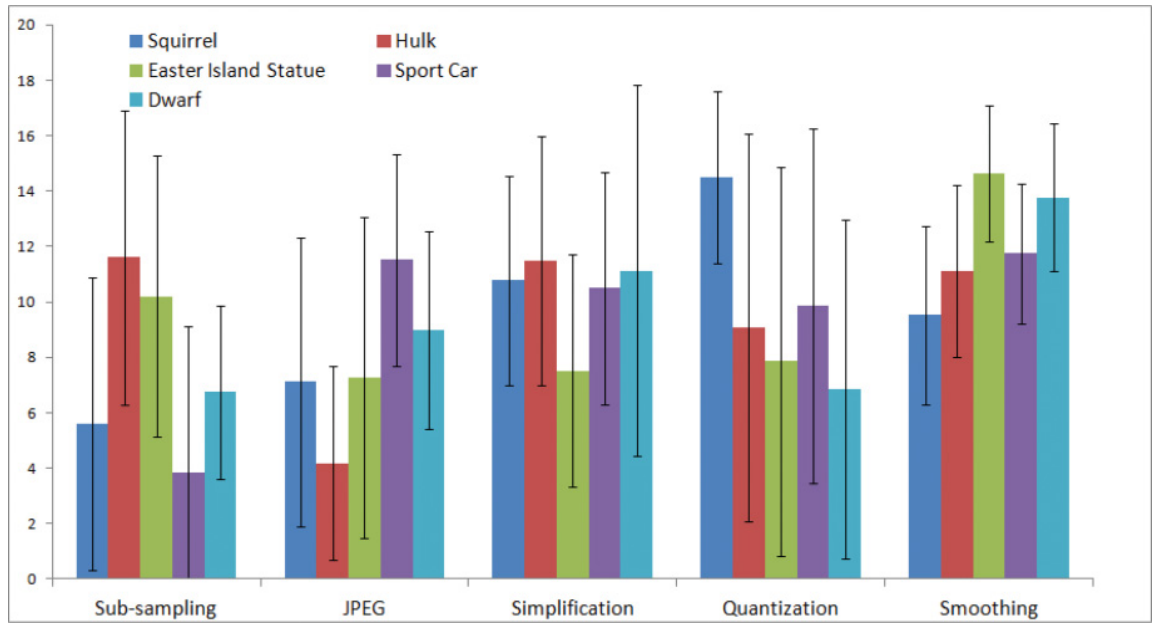

Fig. 4. Mean scores (averaged among the four strengths) for the five types of distortions and the five models for the rendering with shading. Higher scores mean better visual quality.

We also observe that meshes with complex texture seams, like the Statue, are particularly sensitive to simplification, which tends to damage these seams and thus leads to local texture shifting on the surface of the model.

Another observation is that highly curved models (e.g., the Hulk and Squirrel) are very sensitive to Laplacian smoothing which causes large modifications of their shape. Smoother and higher density shapes (e.g., the Dwarf and Statue) remain virtually unchanged by these distortions.

For textures, the effects depend on several factors: their amount of structure, their amount of noise (which relates to the Masking effect), their frequency (which relates to the CSF), and their resolution. For the Sport Car, another important factor is that many of its textures are interior and thus hidden from the observers.

3.6.3 *Influence of the Rendering.* In this section, we investigate the influence of rendering on the visual impact of the distortions. Figure 5 presents the quality scores, averaged over the models for the two rendering conditions: with and without shading. As expected, when only the diffuse albedo of the surface is taken into account, the quality scores of the geometric distortions are consistently better since the impact of the geometry on the rendering is basically limited to the silhouette. By conducting one-tailed paired t-tests, we found a significant increase in the quality of geometric distortions (p-value = $3.9 \times 10^{-3}$) and a significant decrease in the quality of texture distortions (p-value = $1.9 \times 10^{-5}$). These intuitive results have to be kept in mind when assessing the quality of textured 3D models. A calibration according to the rendering may be necessary.

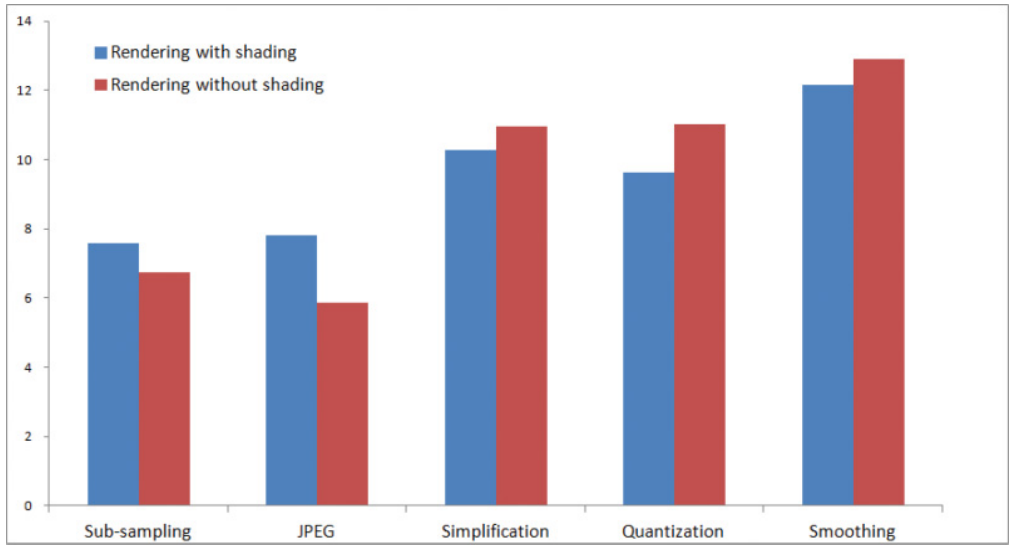

Fig. 5. Mean scores (averaged among the five types of distortions and the four models) for the two types of rendering: with shading (top-left light source, diffuse material) and without shading (diffuse albedo only).

## 4. TOWARD AN OPTIMAL METRIC FOR ASSESSMENT OF TEXTURED MESH QUALITY

In this section, we propose an objective metric for textured mesh quality assessment as a simple linear combination of mesh quality and texture quality. We use the subjective dataset presented above to evaluate the performance of different mesh and texture metrics for this task. We also compare their performance to video metrics computed on the rendered videos. In the next section, we also propose a new metric for geometry quality assessment.

### 4.1 A New Metric for Geometry Quality Assessment

By analyzing the subjective scores obtained for geometric distortions (smoothing, quantization, and simplification), we observed that the ranks of the distortions with the highest strengths from these three types show a pattern common to all models: the distorted model ranked as the worst visual quality always comes from either the strongest quantization method or from the strongest simplification. Distorted models from the strongest smoothing never appear at the end of the ranks. For instance, among the 20 distorted Hulk models, the distortion with the worst visual quality is the 7-bit quantization (subjective score: 0.91), while the distortion of the worst smoothing has a fairly high subjective score (7.45) (see Figure 6). Quality scores from L3,L4 (see Table II for meaning) are actually significantly higher than Q3,Q4 (p-values = 0.011 and 0.0063 for shaded/non-shaded data) and significantly higher than Si3,Si4 (p-values = 0.011 and 0.03). This subjective pattern is related to the perceptual mechanisms of the HVS, which is more sensitive to high-frequency variations on local areas (e.g., distortions caused by simplification or quantization) rather than to more global low-frequency variations (e.g., caused by smoothing).

Based on these observations as well as previous studies which emphasize the reliability of curvature for predicting visual distortions [Lavoué 2011; Wang et al. 2012; Torkhani et al. 2012; Guo et al. 2015], we propose a novel local distortion measurement by computing the variance of curvature differences in local corresponding neighborhoods between two meshes (a distorted mesh $M_d$ and a reference mesh $M_r$). We first establish a correspondence between $M_d$ and $M_r$ (as in Lavoué [2011]) before computing the mean curvature $C$ on each vertex of $M_d$ and its corresponding curvature value $\hat{C}$ on the corresponding point of $M_r$. For each vertex $v$ from $M_d$, we then compute the standard deviation $\delta_v^h$ of the local curvature differences in a connected Euclidean neighborhood of size $h$ around $v$. In a similar way to the Mesh Structural Distortion Measure (MSDM2) [Lavoué 2011], local curvature differences are normalized by the maximum value. $\delta_v^h$ is computed as follows:

$$\delta_v^h = \sqrt{\frac{1}{k}\sum_{j=1}^{k}\left[\frac{\hat{C}_j - C_j}{max(\hat{C}_j, C_j) + a} - E\left(\frac{\hat{C}_j - C_j}{max(\hat{C}_j, C_j) + a}\right)\right]^2}, \quad (2)$$

where $k$ is the number of vertices of the neighborhood, $a$ is a constant to avoid instability when denominators are close to zero, and $E(\frac{\hat{C}_j - C_j}{max(\hat{C}_j, C_j)+a})$ is the mean value of curvature differences in the neighborhood. $\delta_v^h$ is theoretically upper bounded by 2 but can be clamped in [0,1]. Still, as in Lavoué [2011], we carry out this computation for three different neighborhood sizes $h_i$ to capture the perceptually meaningful scales and improve the efficiency and robustness of the metric. We took three scales $h_i \in \{2\epsilon, 3\epsilon, 4\epsilon\}$, where $\epsilon = 2.5\%$ of the max length of the bounding box of the model. Then, the multi-scale local distortion measurement $\delta_M(v)$ is computed as: $\delta_M(v) = \frac{1}{n}\sum_{i=1}^{n}\delta_v^{h_i}$. $n$ is the number of scales (three in our study). Finally, we consider a root mean square pooling of these local measurements to obtain our global Standard Deviation of Curvature Difference SDCD quality index:

$$SDCD(M_d, M_r) = 1 - \left(\frac{1}{|M_d|}\sum_{v \in M_d}\delta_M(v)^2\right)^{\frac{1}{2}}. \quad (3)$$

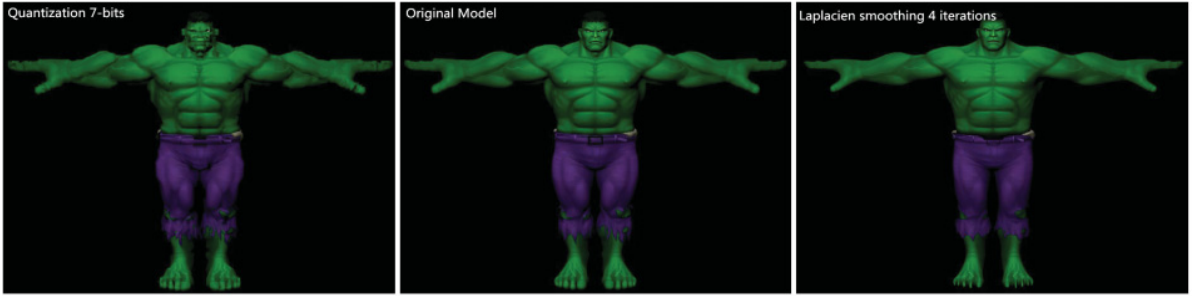

Fig. 6. From left to right: Hulk model with 7-bit quantization (subjective score: 0.91), Original Hulk model, and Hulk model with Laplacian smoothing of four iterations (subjective score: 7.45).

SDCD is within the range [0, 1]: a value of 0 means that the two objects are identical while values near 1 mean that they are visually very different. This metric captures local roughness variations, whereas more global changes (e.g., global shrinking of the model) are not considered as perceptually significant. The performance of this metric is evaluated in the following section.

### 4.2 Mesh and Image Metric Evaluation

The distorted models from our dataset are associated with a single attack (either on geometry or texture). Their subjective scores may thus be used as the ground truth to evaluate respectively mesh metrics and texture image metrics separately. The objective is to determine the most appropriate metrics for geometry and texture image, respectively, and then to combine them (see next section). For each of our reference models, we split the dataset into two groups according to the distortion type: geometry or texture. We selected several commonly used perceptual geometric and image metrics. For geometry, we selected the Root Mean Square Error (RMSE) computed on geometry, MSDM2 [Lavoué 2011], which is one of the best performing perceptually-motivated metrics, and our newly proposed metric SDCD. For texture, we selected the RMSE on image pixels, SSIM [Wang et al. 2004], and MS-SSIM [Wang et al. 2003] (top performing metrics on natural images). Tables IV and V detail the Spearman and Pearson correlations between the objective metrics and the subjective scores for geometry and texture quality assessment, respectively.

Table IV. Performance Comparison (Pearson $r_p$ and Spearman $r_p$ Correlations) of Several Geometric Metrics on Our Subjective Database (Geometric Distortions Only). Rendering with Shading

|  | Squirrel | | Hulk | | Statue | | Sport Car | | Dwarf | | Average | |
| --- | --- | --- | --- | --- | --- | --- | --- | --- | --- | --- | --- | --- |
|  | $r_p$ | $r_s$ | $r_p$ | $r_s$ | $r_p$ | $r_s$ | $r_p$ | $r_s$ | $r_p$ | $r_s$ | $r_p$ | $r_s$ |
| **Geometry RMSE** | 0.72 | **0.90** | 0.37 | 0.61 | 0.07 | 0.00 | **0.65** | 0.28 | 0.02 | 0.28 | 0.37 | 0.41 |
| **MSDM2** [Lavoué 2011] | **0.78** | 0.76 | **0.88** | **0.85** | 0.15 | 0.15 | 0.53 | **0.52** | 0.86 | **0.90** | 0.64 | 0.64 |
| **SDCD** (Our Metric) | 0.60 | 0.55 | 0.84 | 0.80 | **0.64** | **0.64** | 0.40 | 0.41 | **0.89** | 0.86 | **0.67** | **0.65** |

Table V. Performance Comparison (Pearson $r_p$ and Spearman $r_p$ Correlations) of Several Image Metrics on Our Subjective Database (Texture Distortions Only). Rendering with Shading

|  | Squirrel | | Hulk | | Statue | | Sport Car | | Dwarf | | Average | |
| --- | --- | --- | --- | --- | --- | --- | --- | --- | --- | --- | --- | --- |
|  | $r_p$ | $r_s$ | $r_p$ | $r_s$ | $r_p$ | $r_s$ | $r_p$ | $r_s$ | $r_p$ | $r_s$ | $r_p$ | $r_s$ |
| **Image RMSE** | 0.73 | 0.86 | 0.21 | 0.28 | 0.39 | 0.50 | **0.93** | **1.00** | 0.65 | 0.73 | 0.58 | 0.67 |
| **SSIM** [Wang et al. 2004] | 0.27 | 0.17 | 0.60 | 0.77 | 0.70 | 0.69 | 0.90 | 0.93 | 0.73 | 0.88 | 0.64 | 0.69 |
| **MS-SSIM** [Wang et al. 2003] | **0.83** | **0.98** | **0.67** | **0.81** | **0.70** | **0.76** | 0.86 | 0.86 | **0.86** | **0.95** | **0.78** | **0.87** |

Previous studies dedicated to quality assessment of 3D meshes [Corsini et al. 2013] and natural images [Zhang 2012] have shown that geometry and image RMSEs are not good predictors of visual quality. It is interesting to observe that these results are confirmed for our database for which they are outperformed by perceptual metrics. This result is interesting because, in our case, texture and geometry involve complex masking effects. Indeed, texture artifacts may be masked by geometric mapping and vice versa.

Table IV shows that, on certain models, MSDM2 performs better than SDCD. However, for the *Statue*, SDCD demonstrates a significant improvement. Considering these performances, we will consider both MSDM2 and SDCD for our newly combined quality metric for textured mesh. For the texture metric (see Table V), MS-SSIM provides the best overall performance and will thus be chosen for our novel optimal combination.

### 4.3 Toward an Optimal Combination

We propose assessing the visual quality of a textured mesh as a simple linear combination of its geometry quality and its texture quality, measured, respectively, by a 3D mesh metric $Q_G$ (MSDM2 or SDCD) and an image metric $Q_T$ (MS-SSIM). The results show that this simple scheme can provide very good results. Our combined metric is thus defined as follows:

$$CM = \alpha Q_G + (1 - \alpha) Q_T, \qquad (4)$$

where $\alpha$ is an optimal weight determined by greedy optimization through a five-fold cross-validation. For each model (respectively, Squirrel, Statue, Sport Car, Hulk, and Dwarf), we compute the optimal weight as that which maximizes the Spearman correlation over the four other models. We introduce two versions of our combined metric: $CM_1$ (the optimal combination of MSDM2 and MS-SSIM) and $CM_2$ (the optimal combination of SDCD and MS-SSIM). Note that $Q_T$ and $Q_G$, and thus $CM_1$ and $CM_2$, are *similarity* indices, ranging from 0 (total dissimilarity) to 1 (perfect similarity). Table VI details the computed $\alpha$ values.

Table VI. $\alpha$ Values for Each Model and Each Rendering Setting

|  | Squirrel | | Hulk | | Statue | | Sport Car | | Dwarf | |
|---|---|---|---|---|---|---|---|---|---|---|
|  | $\alpha CM_1$ | $\alpha CM_2$ | $\alpha CM_1$ | $\alpha CM_2$ | $\alpha CM_1$ | $\alpha CM_2$ | $\alpha CM_1$ | $\alpha CM_2$ | $\alpha CM_1$ | $\alpha CM_2$ |
| **With shading** | 0.086 | 0.117 | 0.108 | 0.184 | 0.103 | 0.133 | 0.086 | 0.132 | 0.061 | 0.111 |
| **Without shading** | 0.061 | 0.109 | 0.026 | 0.108 | 0.061 | 0.118 | 0.061 | 0.109 | 0.061 | 0.109 |

Table VII. Performance Comparison (Pearson $r_p$ and Spearman $r_p$ Correlations, RMSE of the Residuals) of Several Textured Mesh Quality Metrics on Our Subjective Database. Rendering with Shading

|  | Squirrel | | | Hulk | | | Statue | | | Sport Car | | | Dwarf | | |
|---|---|---|---|---|---|---|---|---|---|---|---|---|---|---|---|
|  | $r_p$ | $r_s$ | RMS | $r_p$ | $r_s$ | RMS | $r_p$ | $r_s$ | RMS | $r_p$ | $r_s$ | RMS | $r_p$ | $r_s$ | RMS |
| **Video-DCT** | 0.12 | 0.09 | 4.76 | 0.18 | 0.36 | 4.99 | 0.26 | 0.30 | 5.04 | 0.62 | 0.67 | 3.97 | 0.24 | 0.25 | 4.75 |
| **Video-PSNR** | 0.22 | 0.26 | 4.68 | 0.33 | 0.36 | 4.79 | 0.21 | 0.26 | 5.12 | 0.67 | 0.70 | 3.68 | 0.31 | 0.32 | 4.63 |
| **Video-MS-SSIM** | 0.24 | 0.39 | 4.64 | 0.17 | 0.41 | 5.00 | 0.25 | 0.42 | 5.05 | 0.67 | **0.72** | 3.64 | 0.38 | 0.40 | 4.50 |
| **FQM** | 0.80 | **0.85** | 2.83 | 0.41 | 0.56 | 4.55 | 0.26 | 0.18 | 5.07 | 0.67 | 0.47 | 3.86 | 0.38 | 0.41 | 4.53 |
| $CM_1$ **(Our Metric)** | **0.82** | 0.82 | **2.24** | **0.81** | 0.81 | **2.92** | 0.36 | 0.30 | 4.93 | **0.70** | 0.60 | **3.58** | 0.56 | 0.70 | 3.95 |
| $CM_2$ **(Our Metric)** | 0.78 | 0.72 | 2.80 | 0.73 | 0.72 | 3.48 | **0.62** | **0.68** | **3.94** | 0.68 | 0.51 | 3.72 | **0.58** | 0.70 | **3.87** |

4.4 Performance Evaluation and Comparison

We compare our metrics $CM_1$ and $CM_2$ to several state-of-the-art metrics:

—FQM [Tian and AlRegib 2004, 2008], a metric especially designed for textured mesh quality assessment. It is defined as a weighted combination of two simple error measurements: the mean squared surface distance and the mean squared error over texture pixels. Optimal weights are computed using cross-validation, as for our metric.
—Several video quality metrics applied to the rendered videos: The Discrete Cosinus Transform-based video quality metric from Xiao [2000], the PSNR applied on all frames and averaged, and the MS-SSIM applied on all frames and averaged. These metrics were computed using the MSU Video Quality Measurement Tool[1].

The performance of these metrics is evaluated using the Spearman and Pearson correlations between the objective metric values and the subjective scores, as well as the RMS error. The Pearson correlation and the RMS are computed after a logistic regression which provides a non-linear mapping between the objective and subjective scores. Results are shown in Tables VII and VIII for the renderings with and without shading, respectively. Scatter plots of subjective scores versus metric values are presented in Figure 7 (shaded rendering).

As illustrated in the tables, our metrics $CM_1$ and $CM_2$ outperform the others for most of the models. Given the fact that, for a given model, the weighting factor $\alpha$ is learned using the other ones, this good performance demonstrates an excellent inter-model robustness. It is interesting to see that the performance of our metrics is generally better for the non-shaded rendering than for the shaded rendering. The reason is that shading involves complex masking interactions between texture and geometry that are not considered in our metrics since they evaluate geometry and texture separately. These interactions are extremely limited in the non-shaded rendering, which accounts for the improved results of our metrics.

---

[1]http://compression.ru/video/quality_measure/video_measurement_tool_en.html.

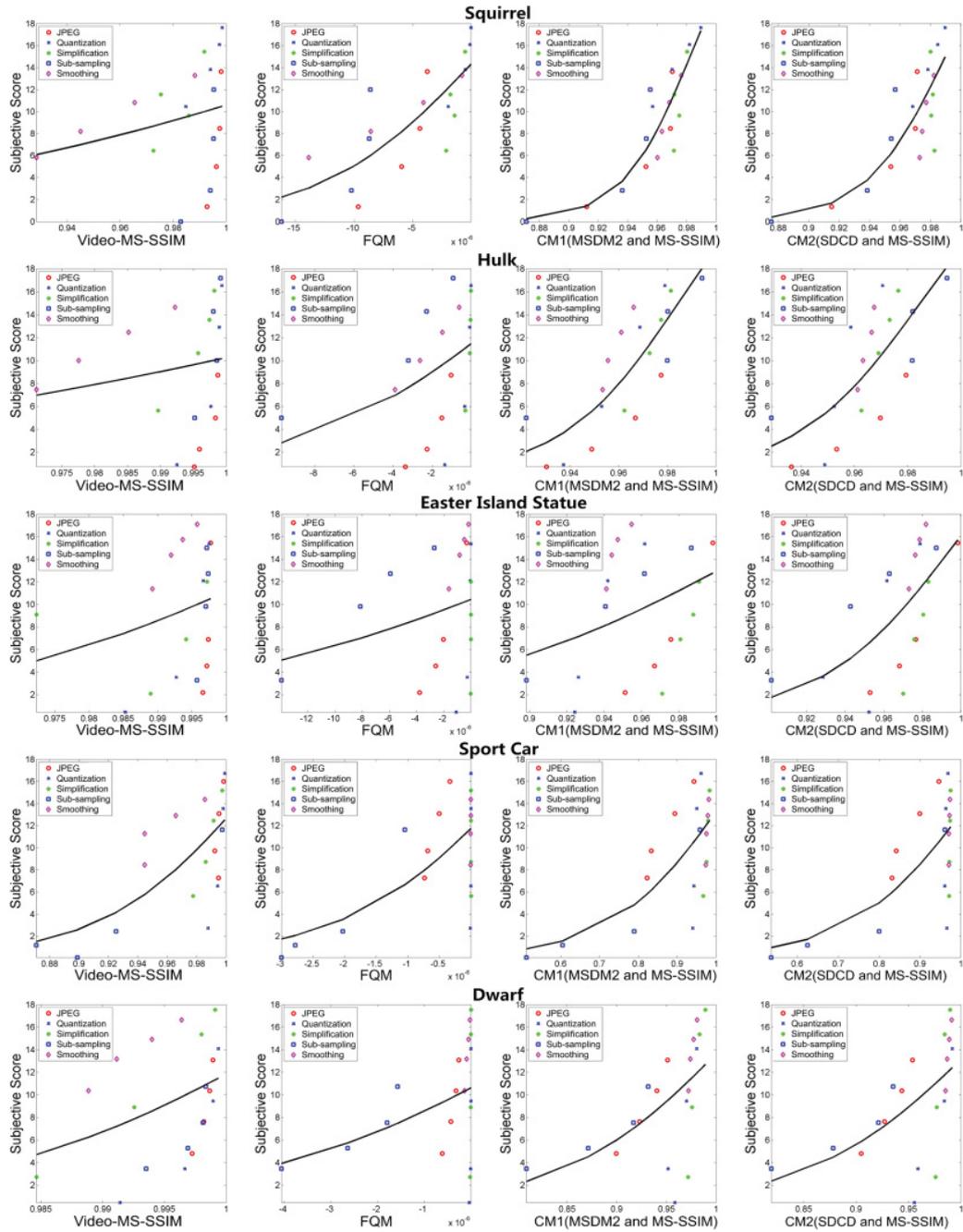

Fig. 7. Scatter plots of subjective scores versus objective metric values. Each point represents one distorted model. Fitted logistic curves are represented in black.

Table VIII. Performance Comparison (Pearson $r_p$ and Spearman $r_p$ Correlations, RMSE of the Residuals) of Several Textured Mesh Quality Metrics on Our Subjective Database. Rendering without Shading

|  | Squirrel | | | Hulk | | | Statue | | | Sport Car | | | Dwarf | | |
|---|---|---|---|---|---|---|---|---|---|---|---|---|---|---|---|
|  | $r_p$ | $r_s$ | RMS | $r_p$ | $r_s$ | RMS | $r_p$ | $r_s$ | RMS | $r_p$ | $r_s$ | RMS | $r_p$ | $r_s$ | RMS |
| **Video-DCT** | −0.21 | −0.26 | 4.76 | 0.22 | 0.25 | 4.72 | 0.34 | 0.24 | 4.74 | 0.58 | 0.42 | 4.23 | 0.24 | 0.21 | 4.89 |
| **Video-PSNR** | −0.13 | −0.04 | 4.82 | 0.35 | 0.39 | 4.50 | 0.13 | 0.22 | 4.97 | 0.51 | 0.56 | 4.32 | 0.31 | 0.29 | 4.78 |
| **Video-MS-SSIM** | −0.05 | 0.14 | 4.86 | 0.28 | 0.61 | 4.64 | 0.34 | 0.37 | 4.72 | 0.67 | 0.69 | 3.77 | 0.65 | 0.63 | 3.80 |
| **FQM** | 0.69 | 0.71 | 3.47 | 0.46 | 0.62 | 4.17 | 0.48 | 0.31 | 4.44 | 0.73 | 0.82 | 3.34 | 0.56 | 0.69 | 4.15 |
| **$CM_1$ (Our Metric)** | **0.80** | **0.88** | **2.38** | 0.69 | 0.80 | 3.29 | 0.47 | 0.41 | 4.53 | **0.77** | **0.90** | **2.86** | 0.73 | 0.81 | 3.23 |
| **$CM_2$ (Our Metric)** | 0.79 | 0.81 | 2.54 | **0.72** | **0.82** | **2.93** | **0.63** | **0.69** | **3.81** | 0.74 | 0.88 | 3.12 | **0.76** | **0.88** | **3.04** |

The main problem with video-based metrics is that they overestimate the visual effect of the Laplacian smoothing, as can be seen in the first column of Figure 7. Indeed, smoothing tends to produce slight displacements of the object silhouette. While this effect is almost invisible to the human eye, it produces displacements of the salient edges in the rendered images, which are very harmful for image/video metrics.

All combined metrics ($CM_1$, $CM_2$, and FQM) have lower performances for the Statue. The reason behind this is visible in Figure 7. Indeed, in the corresponding plots, we observe that the visual impact of simplification distortions (green dots) is underestimated by these metrics (which provide good quality scores for simplified models). The reason is that the poor subjective quality of these distorted versions is due to the damage on the texture seams, which considerably alters the visual appearance but is totally unpredictable by combined metrics that do not take into account the texture mapping. The same underestimation of the simplification impact is observed for the Dwarf model, which also exhibits complex texture seams. For this latter model, the geometric metrics also underestimate the impact of quantization which is particularly harmful for such a high-resolution model.

One last observation is the superiority of video-based metrics for the Sport Car model (shaded rendering). The reason is as follows: this model has several interior parts (e.g., seats, radio) whose texture maps are severely damaged by JPEG and sub-sampling distortions. However, these interior parts are almost invisible due to the position of the camera in the videos. Hence, the subjective scores are rather good for these distorted models, whereas the image metrics predict very low-quality values. On the contrary, video-based metrics only take into account the visible parts and thus provide correct results (in particular Video-MS-SSIM). To verify this effect, we adopted a slightly different viewpoint for this particular model, in the non-shaded rendering, which improves the visibility of these interior parts. As expected, the results of the combined metrics are much better (see Table VIII). This observation argues for the integration of a visibility information in the combined metrics as an efficient way to overcome this drawback.

### 4.5 Validation on Compound Distortions

In this section, we validate our objective metrics using a new set of compound geometry-texture distortions. We selected the Dwarf model, and we manually selected 36 distorted versions among the $12 \times 8 = 96$ possible combinations of the 12 geometry and 8 texture distortions detailed in Section 3.1, resulting in a new validation set of 36 models. Details about these mixed distortions are available in the supplementary material; some examples are shown in Figure 2, bottom right. As before, we created two sets of videos (each of 10 seconds duration) with and without shading, respectively. We performed a paired-comparison experiment to obtain the subjective scores. For this mixed-distortion setting, the hypothesis $Q_{D_i} > Q_{D_j}, \forall j > i$ from our sorting algorithm no longer holds. Hence, we implemented a more classical self-balancing binary tree, as in Mantiuk et al. [2012]. The average number of comparisons for sorting the 36 distorted models was 140 (instead of 630 for the full comparison). Twenty observers took part in this new experiment, 19 rated 1 set and 1 rated both sets. We thus obtained 10 judgments for the shaded rendering and 11 for the non-shaded one. The average time to finish the experiment for one video set of 36 videos was 21 minutes. As for the previous experiment, the agreement is rather high (respectively, 0.75 and 0.71 for shaded and non-shaded settings). Detailed raw scores are included in the supplementary material.

Table IX. Performance Comparison on the Compound Distortion Dataset

|  | With shading | | | Without shading | | |
|---|---|---|---|---|---|---|
|  | $r_p$ | $r_s$ | RMS | $r_p$ | $r_s$ | RMS |
| **Video-DCT** | 0.32 | 0.50 | 7.81 | 0.35 | 0.32 | 8.23 |
| **Video-PSNR** | 0.33 | 0.58 | 7.32 | 0.40 | 0.33 | 8.05 |
| **Video-MS-SSIM** | 0.67 | 0.66 | 6.79 | 0.68 | 0.67 | 6.42 |
| **FQM** | 0.64 | 0.66 | 6.90 | 0.76 | 0.77 | 5.73 |
| $CM_1$ (Our Metric) | 0.74 | 0.77 | 6.01 | 0.85 | 0.86 | 4.47 |
| $CM_2$ (Our Metric) | **0.80** | **0.85** | **5.33** | **0.86** | **0.87** | **4.44** |

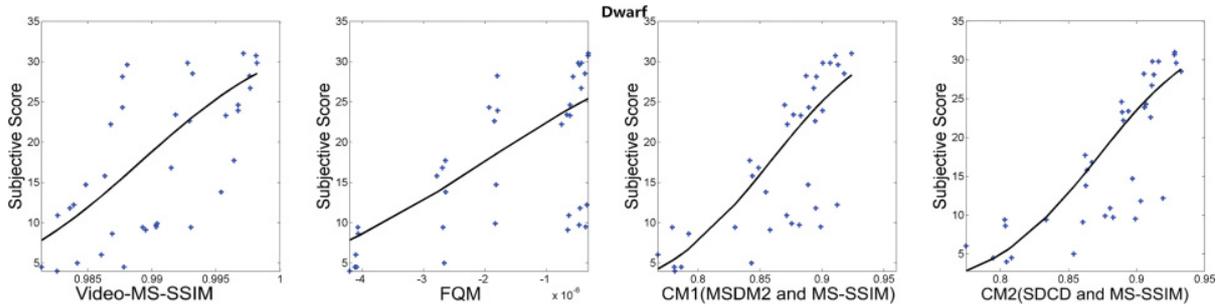

Fig. 8. Scatter plots of subjective scores versus objective metric values for the compound distortion dataset. Each point represents one distorted model. Fitted logistic curves are represented in black.

To validate our metrics, we computed their optimal weights based on the models from our previous experiment (we excluded the Dwarf). Results are detailed in Table IX and Figure 8. Even for this difficult scenario (learning on single-type distortions and testing on compound distortions), our metrics offer excellent performance: 0.85 and 0.87 Spearman correlations for $CM_2$, for both rendering settings, hence, demonstrating once again an excellent robustness.

## 5. CONCLUSION AND PERSPECTIVES

In this work, we designed and constructed a new subjectively-rated database of textured 3D meshes. Our subjective study is based on a paired comparison protocol and involved more than 100 subjects. The database contains 136 distorted models (subject to geometry and texture distortions), which were evaluated within two rendering settings. The subjective results allowed us to draw interesting conclusions regarding the influence of the shape and texture content as well as the rendering on the perceptual impact of distortions. We then proposed new objective metrics for visual quality assessment of textured meshes as optimized linear combinations of mesh quality and texture quality. We used our subjective dataset to evaluate the performance of these metrics against state-of-the-art ones and explained the failure of metrics applied on the rendered images/videos. Such perceptually-validated metrics are of great interest for many applications such as 3D model simplification or compression, texture simplification, and so on. We also proposed a new measure for geometry quality assessment. Note that our dataset, subjective scores, and metric results will be made publicly available on-line. While our proposed metrics showed that they outperformed their counterparts for the task of textured mesh quality assessment, there is still room for improvement. For instance, when evaluating geometry and texture quality, we need to integrate some visibility information. Indeed, the interior parts of a 3D model do not contribute to its rendered visual appearance. As another example, regions in very convex areas will only be visible from a few viewpoints and thus will have little impact on subjective opinion. Another issue is the variation in texel size within a texture; indeed, different regions of a texture map are not necessarily mapped with the same size on the screen, and this effect should be taken into account. The angular resolution (in pixels per degree of visual angle) of the rendered scene is also important and may be integrated as a scale factor in the metric. Finally, a major issue is to incorporate texture coordinate distortions which are a common side effect of geometric changes. Based on our observation, even slight movements of texture seams may seriously harm visual appearance. Finally, we could conduct a comprehensive evaluation to determine, among the dozens of existing image metrics, the one best adapted to texture evaluation.

ACKNOWLEDGMENTS

We thank Massimiliano Corsini and the Visual Computing Laboratory of ISTI-CNR for the Dwarf model, as well as Mark Pauly and the EPFL Computer Graphics and Geometry Laboratory for the Squirrel and Statue models. This work is supported in part by the China Scholarship Council and Alberta Innovates Techology Futures.

REFERENCES

Tunç Ozan Aydın, Martin Čadík, Karol Myszkowski, and Hans-Peter Seidel. 2010. Video quality assessment for computer graphics applications. *ACM Transactions on Graphics* 29, 6 (Dec 2010), 1. DOI:http://dx.doi.org/10.1145/1882261.1866187

Martin Čadík, Robert Herzog, Rafal Mantiuk, Radosaw Mantiuk, Karol Myszkowski, and Hans-Peter Seidel. 2013. Learning to predict localized distortions in rendered images. In *Pacific Graphics*, Vol. 32.

Martin Čadík, Robert Herzog, Rafal Mantiuk, Karol Myszkowski, and Hans-Peter Seidel. 2012. New measurements reveal weaknesses of image quality metrics in evaluating graphics artifacts. *ACM Transactions on Graphics* 31, 6 (2012), Article 147.

D. M. Chandler and S. S. Hemami. 2007. VSNR: A wavelet-based visual signal-to-noise ratio for natural images. *IEEE Transactions on Image Processing* 16, 9 (Sep 2007), 2284–2298.

I. Cheng and A. Basu. 2007. Perceptually optimized 3-D transmission over wireless networks. *IEEE Transactions on Multimedia* 9, 2 (Feb 2007), 386–396. DOI:http://dx.doi.org/10.1109/TMM.2006.886291

Massimiliano Corsini, Elisa Drelie Gelasca, Touradj Ebrahimi, and Mauro Barni. 2007. Watermarked 3-D mesh quality assessment. *IEEE Transactions on Multimedia* 9, 2 (Feb 2007), 247–256.

M. Corsini, M. C. Larabi, G. Lavoué, O. Petrik, L. Váša, and K. Wang. 2013. Perceptual metrics for static and dynamic triangle meshes. *Computer Graphics Forum* 32, 1 (Feb 2013), 101–125.

Scott Daly. 1993. The visible differences predictor: An algorithm for the assessment of image fidelity. In *Digital Images and Human Vision*, Andrew B. Watson (Ed.). MIT Press, Cambridge, 179–206.

J. Ferwerda, S. Pattanaik, P. Shirley, and D. Greenberg. 1997. A model of visual masking for computer graphics. In *ACM SIGGRAPH*. 143–152.

M. Garland and P.-S. Heckbert. 1997. Surface simplification using quadric error metrics. In *ACM SIGGRAPH*. 209–216.

Z. Govindarajulu, M. Kendall, and J. D. Gibbons. 1992. Rank correlation methods. *Technometrics* 34, 1 (Feb 1992), 108. DOI:http://dx.doi.org/10.2307/1269571

Wesley Griffin and Marc Olano. 2015. Evaluating texture compression masking effects using objective image quality assessment metrics. *IEEE Transactions on Visualization and Computer Graphics* 21, 8 (2015), 970–079. DOI:http://dx.doi.org/10.1109/TVCG.2015.2429576

Jinjiang Guo, Vincent Vidal, Atilla Baskurt, and Guillaume Lavoué. 2015. Evaluating the local visibility of geometric artifacts. In *Proceedings of the ACM Symposium in Applied Perception*.

Robert Herzog, Martin Čadík, Tunç O. Aydın, Kwang In Kim, Karol Myszkowski, and Hans-P. Seidel. 2012. NoRM: No-reference image quality metric for realistic image synthesis. *Computer Graphics Forum* 31, 2 (Pt 3), 545–554. DOI:http://dx.doi.org/10.1111/j.1467-8659.2012.03055.x

Z. Karni and C. Gotsman. 2000. Spectral compression of mesh geometry. In *ACM Siggraph*. 279–286.

Guillaume Lavoué. 2011. A multiscale metric for 3D mesh visual quality assessment. *Computer Graphics Forum* 30, 5 (2011), 1427–1437.

G. Lavoué, M. C. Larabi, and Libor Váša. 2016. On the efficiency of image metrics for evaluating the visual quality of 3D models. *IEEE Transactions on Visualization and Computer Graphics* 22, 8 (2016), 1987–1999. DOI:http://dx.doi.org/10.1109/TVCG.2015.2480079

Guillaume Lavoué and Rafa Mantiuk. 2015. Quality assessment in computer graphics. *Visual Signal Quality Assessment: Quality of Experience (QoE)* (2015), 243–286. DOI:http://dx.doi.org/10.1007/978-3-319-10368-6_9

Patrick Ledda, Alan Chalmers, Tom Troscianko, and Helge Seetzen. 2005. Evaluation of tone mapping operators using a high dynamic range display. *ACM Transactions on Graphics* 24, 3 (Jul 2005), 640.

Jeffrey Lubin. 1993. The use of psychophysical data and models in the analysis of display system performance. In *Digital Images and Human Vision*, A. B. Watson (Ed.). 163–178.

J. Mannos and D. Sakrison. 1974. The effects of a visual fidelity criterion of the encoding of images. *IEEE Transactions on Information Theory* 20, 4 (Jul 1974), 525–536. DOI:http://dx.doi.org/10.1109/tit.1974.1055250

Rafal Mantiuk, Kil Joong Kim, Allan G. Rempel, and Wolfgang Heidrich. 2011. HDR-VDP-2: A calibrated visual metric for visibility and quality predictions in all luminance conditions. In *ACM Transactions on Graphics (Proc. of SIGGRAPH'11)* 30, 4 (2011), Article no. 40.

Rafa K. Mantiuk, Anna Tomaszewska, and Radosaw Mantiuk. 2012. Comparison of four subjective methods for image quality assessment. *Computer Graphics Forum* 31, 8 (Dec 2012), 2478–2491. DOI:http://dx.doi.org/10.1111/j.1467-8659.2012.03188.x

Georges Nader, Kai Wang, H. Franck, and Florent Dupont. 2016. Just noticeable distortion profile for flat-shaded 3D mesh surfaces. *IEEE Transactions on Visualization and Computer Graphics* (2016). DOI:http://dx.doi.org/10.1109/TVCG.2015.2507578

J. P. O'Shea, M. S. Banks, and M. Agrawala. 2008. The assumed light direction for perceiving shape from shading. In *Proceedings of the Symposium on Applied Perception in Graphics and Visualization*.

Yixin Pan, Irene Cheng, and Anup Basu. 2005. Quality metric for approximating subjective evaluation of 3-D objects. *IEEE Transactions on Multimedia* 7, 2 (Apr 2005), 269–279.

Lijun Qu and G. W. Meyer. 2008. Perceptually guided polygon reduction. *IEEE Transactions on Visualization and Computer Graphics* 14, 5 (2008), 1015–1029. DOI:http://dx.doi.org/10.1109/TVCG.2008.51

Bernice E. Rogowitz and H. Rushmeier. 2001. Are image quality metrics adequate to evaluate the quality of geometric objects? In *Proceedings of SPIE*. 340–348.

Kalpana Seshadrinathan, Rajiv Soundararajan, Alan Conrad Bovik, and Lawrence K. Cormack. 2010. Study of subjective and objective quality assessment of video. *IEEE Transactions on Image Processing* 19, 6 (Jun 2010), 1427–41. DOI:http://dx.doi.org/10.1109/TIP.2010.2042111

H. R. Sheikh and A. C. Bovik. 2006. Image information and visual quality. *IEEE Transactions on Image Processing* 15, 2 (Feb 2006), 430–444.

D. Amnon Silverstein and Joyce E. Farrell. 2001. Efficient method for paired comparison. *Journal of Electronic Imaging* 10, 2 (2001), 394. DOI:http://dx.doi.org/10.1117/1.1344187

J. Ström and T. Akenine-Möller. 2005. i PACKMAN: High-quality, low-complexity texture compression for mobile phones. In *Proceedings of the ACM SIGGRAPH/EUROGRAPHICS Conference on Graphics Hardware*. 177–182. DOI:http://dx.doi.org/10.1145/1071866.1071877

Jennifer Sun and Pietro Perona. 1998. Where is the sun? *Nature Neuroscience* (1998), 183–184. Retrieved from http://www.nature.com/neuro/journal/v1/n3/abs/nn0798.

G. Taubin. 1995. A signal processing approach to fair surface design. In *ACM Siggraph*. 351–358.

L. L. Thurstone. 1927. A law of comparative judgments. *Psychological Review* 34 (1927), 273–286.

Dihong Tian and Ghassan AlRegib. 2004. FQM. In *Proceedings of the 12th Annual ACM International Conference on Multimedia - MULTIMEDIA 04. Association for Computing Machinery (ACM)*. DOI:http://dx.doi.org/10.1145/1027527.1027684

Dihong Tian and G. AlRegib. 2008. Batex3: Bit allocation for progressive transmission of textured 3-D models. *IEEE Transactions on Circuits and Systems for Video Technology* 18, 1 (2008), 23–35.

Fakhri Torkhani, Kai Wang, and Jean-Marc Chassery. 2012. A curvature tensor distance for mesh visual quality assessment. In *Proceedings of the International Conference on Computer Vision and Graphics*.

Libor Váša and Jan Rus. 2012. Dihedral angle mesh error: A fast perception correlated distortion measure for fixed connectivity triangle meshes. *Computer Graphics Forum* 31, 5 (2012), 1715–1724.

Kai Wang, Fakhri Torkhani, and Annick Montanvert. 2012. A fast roughness-based approach to the assessment of 3D mesh visual quality. *Computers & Graphics* 36, 7 (2012), 808–818.

Z. Wang, A. C. Bovik, H. R. Sheikh, and E. P. Simoncelli. 2004. Image quality assessment: From error visibility to structural similarity. *IEEE Transactions on Image Processing* 13, 4 (Apr 2004), 600–612. DOI:http://dx.doi.org/10.1109/tip.2003.819861

Zhou Wang and Alan C. Bovik. 2006. *Modern Image Quality Assessment*. Vol. 2. Morgan & Claypool Publishers. DOI:http://dx.doi.org/10.2200/S00010ED1V01Y200508IVM003

Zhou Wang and Qiang Li. 2011. Information content weighting for perceptual image quality assessment. *IEEE Transactions on Image Processing* 20, 5 (May 2011), 1185–1198. DOI:http://dx.doi.org/10.1109/tip.2010.2092435

Z. Wang, E. P. Simoncelli, and A. C. Bovik. 2003. Multiscale structural similarity for image quality assessment. *IEEE Asilomar Conference on Signals, Systems and Computers* 2, 1 (2003), 1398–1402. DOI:http://dx.doi.org/10.1109/ACSSC.2003.1292216

Benjamin Watson, Alinda Friedman, and Aaron McGaffey. 2001. Measuring and predicting visual fidelity. In *ACM SIGGRAPH*. 213–220.

Feng Xiao. 2000. *DCT-based Video Quality Evaluation*. Technical Report. Stanford University. Retrieved from http://compression.ru/video/quality.

Sheng Yang, Chao-Hua Lee, and C. C. J. Kuo. 2004. Optimized mesh and texture multiplexing for progressive textured model transmission. In *Proceedings of the ACM Multimedia Conference*. 676–683. DOI:http://dx.doi.org/10.1145/1027527.1027683

Hojatollah Yeganeh and Zhou Wang. 2013. Objective quality assessment of tone-mapped images. *IEEE Transactions on Image Processing* 22, 2 (Feb 2013), 657–67. DOI:http://dx.doi.org/10.1109/TIP.2012.2221725

L. Zhang. 2012. A comprehensive evaluation of full reference image quality assessment algorithms. In *Proceedings of the International Conference on Image Processing (ICIP)*. 1477–1480.